%% file: journal.tex
\documentclass[journal,11pt,twocolumn]{IEEEtran}

\usepackage{amsfonts, amsmath, mathrsfs, amsbsy, amssymb, color, graphicx, stmaryrd, caption}
\usepackage{algorithmic, algorithm, booktabs, subfigure, url, multirow}

\include{def}

\begin{document}
\title{JULIA: Joint Multi-linear and Nonlinear Identification for Tensor Completion}
\author {
	Cheng Qian,
	Kejun Huang,
	Lucas Glass,
	Rakshith S. Srinivasa, and
	Jimeng Sun
	\thanks{Corresponding Author: Cheng Qian (alextoqc@gmail.com).\\
		Cheng Qian, Lucas Glass, and Rakshith S. Srinivasa are with Analytics Center of Excellence, IQVIA Inc., United States. Kejun Huang is with Dept. of Computer and Information Science and Engineering, University of Florida, United States. Jimeng Sun is with Department of Computer Science, University of Illinois at Urbana-Champaign, United States.}
}

\maketitle

\begin{abstract}
	Tensor completion aims at imputing missing entries from a partially observed tensor. Existing tensor completion methods often assume either multi-linear or nonlinear relationships between latent components.
	However, real-world tensors have much more complex patterns where both multi-linear and nonlinear relationships may coexist. In such cases, the existing methods are insufficient to describe the data structure. This paper proposes a \textbf{J}oint m\textbf{U}lti-linear and non\textbf{L}inear \textbf{I}dentific\textbf{A}tion (JULIA) framework for large-scale tensor completion. JULIA unifies the multi-linear and nonlinear tensor completion models with several advantages over the existing methods: 1) Flexible model selection, i.e., it fits a tensor by assigning its values as a combination of multi-linear and nonlinear components; 2) Compatible with existing nonlinear tensor completion methods; 3) Efficient training based on a well-designed alternating optimization approach. Experiments on six real large-scale tensors demonstrate that JULIA outperforms many existing tensor completion algorithms. Furthermore, JULIA can improve the performance of a class of nonlinear tensor completion methods. The results show that in some large-scale tensor completion scenarios, baseline methods with JULIA are able to obtain up to 55\% lower root mean-squared-error and save 67\% computational complexity.
\end{abstract}

\section{Introduction}
Real-world tensors are often incomplete due to limited data access, delay during data collection, loss of information, etc. The problem of imputing missing entries from partially observed tensor samples is known as tensor completion. This problem has many applications such as knowledge graph link prediction \cite{zhou2017tensor}, spatio-temporal traffic prediction \cite{said2021spatiotemporal} and healthcare data completion \cite{chen2020learning}. 

Low-rank tensor completion is a popular approach to solve the underlining tensor completion problem, which estimates the missing values through estimating $N$ latent factor matrices from the observed tensor entries. There are mainly two types of low-rank models. One is the multi-linear model that assumes linear relationships of latent components, e.g., Canonical Polyadic (CP) tensor completion model.
The other is a class of nonlinear models, e.g., those based on deep neural networks (DNN) that have recently attracted much attention. 
The multi-linear model has satisfactory performance in identifying the linear components in tensors. Deep nonlinear methods usually have better performance when the underlying tensor models tend to be highly nonlinear. However, training a deep model is much more expensive than training a multi-linear model, since the former requires more data to learn the network parameters. Moreover, the nonlinear tensor completion methods could overfit the data since they ignore the parsimonious multi-linear CP  structure.

In real-world tensors, data relationships can be both multi-linear and nonlinear. For example, in a spatio-temporal tensor indexed by $\mathrm{location}\times\mathrm{Feature}\times\mathrm{Time}$, where the feature mode consists of COVID-19 cases, deaths and hospitalizations. We know that the COVID-19 deaths and hospitalization directly correlate with the COVID-19 cases, which implies a low-rank multi-linear relationship. However, in the late stage of the COVID-19 pandemic, the transmission patterns of different locations become less correlated due to the inconsistent responses/regulations of local governments, which implies a nonlinear data relationship. Once tensors have multi-linear and nonlinear relationships coexisting, neither a multi-linear model nor a nonlinear model is sufficient to describe the tensors accurately.  

In this paper, we propose a unified framework named as \textbf{J}oint m\textbf{U}lti-linear and non\textbf{L}inear \textbf{I}dentific\textbf{A}tion (JULIA) for large-scale tensor completion. Unlike existing methods, JULIA models a tensor using two types of latent components: $R$ multi-linear components and $F$ nonlinear components. Here, the multi-linear components approximate hidden linearity among the highly correlated data points while the nonlinear components describe the hidden nonlinearity in the tensors, which is modeled through a DNN. An alternating optimization (AO) approach is then developed to train JULIA, where the multi-linear and nonlinear components are trained alternately by fixing one for the other. Compared to existing tensor completion frameworks, JULIA has the following advantages:
\begin{itemize}
	\item JULIA unifies the existing tensor models: When $R=0$, it reduces to a nonlinear tensor completion model; when $F=0$, it reduces to the multi-linear tensors. Therefore, JULIA can perform model selection and identify the number of multi-linear and nonlinear components in tensors by tuning the values of $R$ and $F$, and hence it approximates tensors more accurately. 
	
	\item JULIA can handle very large-scale tensor completion tasks. The AO training method enables to accelerate the learning process and offer better performance.
	
	\item JULIA is compatible with any existing nonlinear tensor completion algorithms, and it improves their performance through the AO approach with fewer parameters.
\end{itemize}
Experimental results on six large-scale tensors showcase the effectiveness of JULIA.

\section{Related Work}
We now review the related works on low-rank tensor completion methods. The CP and Tucker decomposition are the two widely used multi-linear models for tensor completion \cite{Sidiropoulos2017}. Numerous variants have been developed under the CP and Tucker frameworks with applications in image and video inpainting \cite{zhou2017tensor}, healthcare data completion \cite{chen2020learning, wang2015rubik}, graph link prediction \cite{kazemi2018simple, kanatsoulis2021tex}, signal reconstruction \cite{kanatsoulis2019tensor}, spatio-temporal tensor completion \cite{qian2021multi}, etc. Recently, there are also deep nonlinear tensor completion models proposed. NeurTN \cite{chen2020learning} combines tensor algebra and deep neural networks for drug-target-disease interaction prediction. AVOCADO \cite{schreiber2020avocado} employs the Multilayer perceptron (MLP) to learn the nonlinear relationship between factor matrices. Unlike NeurTN and AVOCADO which are MLP based, COSTCO \cite{liu2019costco} models the nonlinear relationship through the local embedding features extracted by two convolutional layers. More recently, Sonkar {et al.} \cite{sonkar2021neptune} proposed NePTuNe, which is a nonlinear Tucker-like method for knowledge graph completion. 


\section{Preliminaries}



\subsection{Multi-linear Tensor Completion}
One representative multi-linear tensor factorization model is the CP completion (CPC) which expresses an \mbox{$N$-way} tensor $\tX \in \bR^{I_1 \times \ldots \times I_N}$ as a sum of \mbox{rank-$1$} components, i.e., a multi-linear latent variable model:
\begin{equation*}
	\tX = 
	\sum_{r=1}^R\prod_{n=1}^{N} \A_n(:, r)
\end{equation*}
where $\A_n \in \bR^{I_n \times R}$ stands for the $n$-th factor matrix, $R$ is the tensor rank indicating the minimum number of components needed to synthesize $\tX$ and $\A_n(:,i)$ is the $i$-th column of $\A_n$.

With the multi-linear assumption, CPC can represent an $N$-way tensor of size $I_1 \times \ldots \times I_N$ using only $(\sum_{n=1}^{N} I_n - 1)\times R$ parameters. The CPC has two appealing properties: 1) It is universal, i.e., every tensor admits a CP model of finite rank; 2) It can identify the true latent factors that synthesize $\tX$ under mild conditions. The two properties make CPC a powerful tool for data analysis, especially when we need model interpretability \cite{Sidiropoulos2017}.

\subsection{Nonlinear Tensor Completion}
Unlike the CPC model, the so-called nonlinear tensor completion model employs a nonlinear function $f$ such that the $(i_1,\ldots,i_N)$-th entry in $\tX$ is expressed as.

\begin{equation*}
	x_{i_1\ldots i_N} = f\big(\{\A_n(i_n,\colon)\}_{n=1}^N;\bth\big),
\end{equation*}
where $\A_n(i,:)$ is the $i$-th row of $\A_n$, $f(\cdot)$ is a function that is not multi-linear and passes the rows of factor matrices through a set of learnable parameters $\bth$ to approximate the tensor entries. 

Recent works have shown that deep learning-based tensor completion models perform better than the classical multi-linear models for large-scale tensor completion \cite{liu2019costco}. However, one drawback of such nonlinear models is the lack of identifiability guarantee and model interpretability.

\begin{figure*}[ht]
	\centering
	\includegraphics[width=0.75\linewidth]{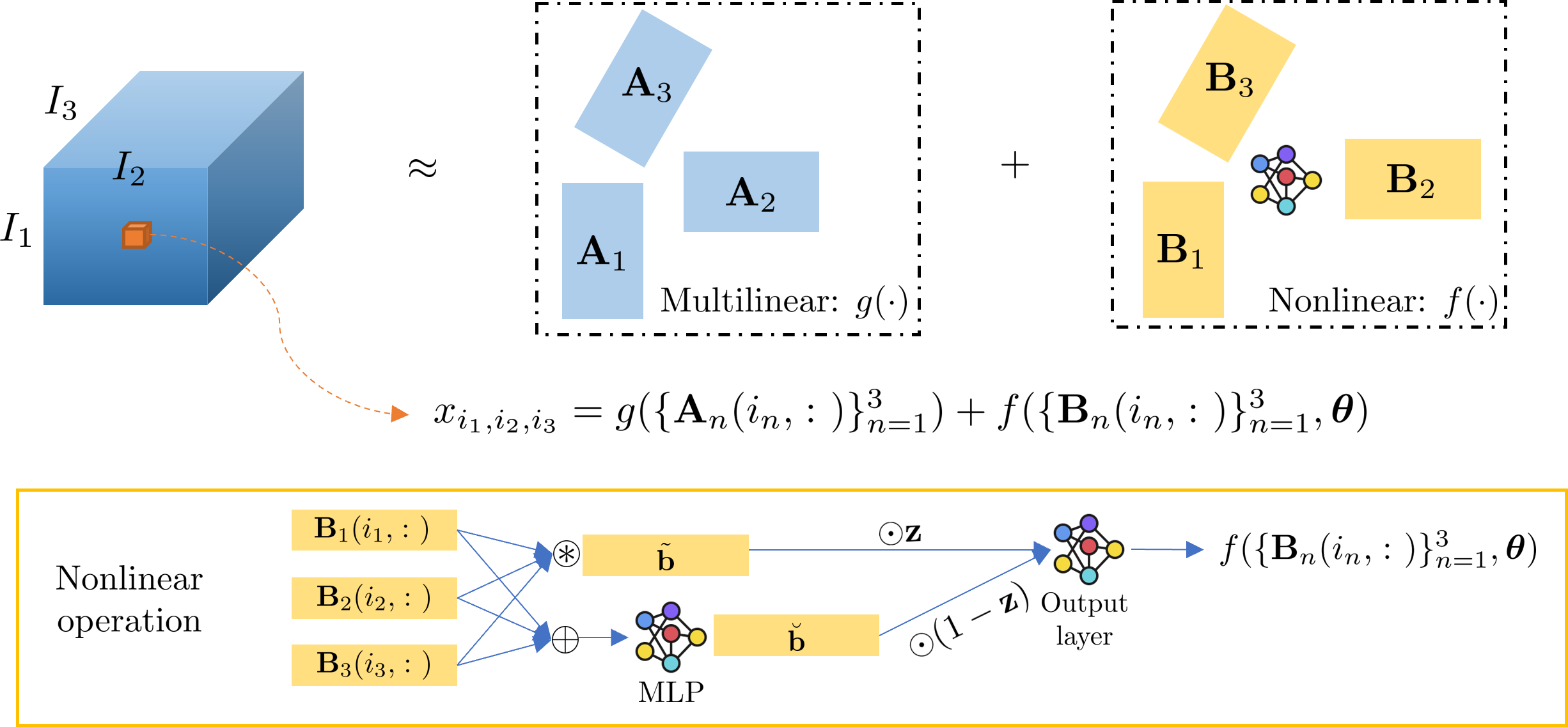}
	\caption{Flowchart of JULIA modeling a 3-way tensor.}
	\label{fig:flow}
\end{figure*}

\section{Proposed Method}
In this section, we present JULIA - an efficient framework for tensor completion. 

\subsection{Motivation}
Due to the complexity of real-world tensors, linear and nonlinear components may coexist in the latent subspace. Unlike many existing tensor methods that use a set of unified latent components operated either linearly or nonlinearly, JULIA has two different sets of factor matrices, i.e., one consists of the multi-linear factors. In contrast, the other one consists of nonlinear factors, to capture the complex patterns in real-world tensors. 

\subsection{Model}
Given an $N$-way tensor $\tX\in\bR^{I_1\times\ldots\times I_N}$, let us define $\{\A_1,\ldots,\A_N\}$ as the multi-linear set which captures the linear relationship in the tensor and each factor matrix has $R$ independent components, i.e., $\A_n = [\a_{n,1}, \ldots, \a_{n, R}]\in\bR^{I_n\times R}$. Similarly, we define $\{\B_1,\ldots,\B_N\}$ as the nonlinear set which captures the nonlinear relationship in the tensor and each factor matrix has $F$ independent components, i.e., $\B_n = [\a_{n,1}, \ldots, \a_{n, F}]\in\bR^{I_N\times F}$.

With the above notations, JULIA models the $(i_1,\ldots,i_N)$-th tensor element as
\begin{align}\label{eq:juliax}
	x_{i_1\ldots i_N} = g\left(\{\A_n(i_n,\colon)\}_{n=1}^N\right) + f\left(\{\B_n(i_n,\colon)\}_{n=1}^N; \bth\right)
\end{align}
where $g$ is the multi-linear function, i.e., 
\begin{align}\label{eq:g}
	g\left(\{\A_n(i_n,\colon)\}_{n=1}^N\right) = \sum_{r=1}^R\prod_{n=1}^{N} \A_n(i_n, r)
\end{align}
and $f(\cdot)$ is the nonlinear function that is parameterized by $\bth$ and takes the rows of the nonlinear factor matrices $\{\B_1,\ldots,\B_N\}$ as input. Fig. \ref{fig:flow} shows an example of how JULIA models a 3-way tensor.

The latent nonlinear representation of the tensor is captured by a deep neural network built upon $N$ factor matrices $\{\B_1,\cdots,\B_N\}$. 
Specifically, the $(i_1,\cdots,i_N)$-th value $x_{i_1\cdots i_N}$ is calculated based on $\{\B_n(i_n,\colon)\}_{n=1}^N$. The nonlinear function $f(\cdot)$ consists of two major flows. The first flow transforms the $N$ row vectors as
\begin{align*}
	\tilde{\b} = \sigma\left(\B_1(i_1,\colon) \odot \cdots\odot \B_N(i_N,\colon)\right) \in\bR^{F}
\end{align*}
where $\odot$ is the element-wise product, $\sigma(\cdot)$ is the activation function. By default, we consider the ReLU function.

The second flow concatenates $\{\B_n(i_n,\colon)\}_{n=1}^N$ together and passes the concatenated vector through a MLP, which produces $\breve{\b}$, i.e.,
\begin{align*}
	\breve{\b} = \mathrm{MLP}\big(\B_1(i_1,\colon) \oplus \cdots \oplus \B_N(i_N,\colon)\big) \in\bR^{F}
\end{align*}
where $\oplus$ denotes the concatenation operator.
Both flows are then summed over a weighting vector $\z\in\bR^F$
\begin{align*}
	\b = \z\odot \tilde{\b} + (1 - \z)\odot \breve{\b}
\end{align*}
which is finally passed through an output layer to estimate the nonlinear term in $x_{i_1\cdots i_N}$ as
\begin{align}\label{eq:f}
	f(\{\B_n(i_n,:)\}_{n=1}^N) = \sigma\left( \w^T\b + \epsilon \right)
\end{align}
where $\epsilon$ denotes the bias and $(\cdot)^T$ is the transpose operator. 

Substituting \eqref{eq:g} and \eqref{eq:f} into \eqref{eq:juliax} yields
\begin{align}
	\hat{x}_{i_1\cdots i_N} = \sum_{r=1}^R\prod_{n=1}^{N} \A_n(i_n, r) + \sigma\left( \w^T\b + \epsilon \right).
\end{align}

\begin{remark}
	
	\begin{itemize}
		\item JULIA has the freedom of choosing the nonlinear function $f(\cdot)$. 
		The nonlinear function $f(\cdot)$ can be any existing nonlinear tensor completion models, e.g., \cite{liu2019costco,schreiber2020avocado,sonkar2021neptune}. 
		
		\item Unlike many existing tensor models that stick to one tensor rank of either `pure' multi-linear or nonlinear components, JULIA has the freedom to change ranks. In \eqref{eq:juliax}, the total number of components is $(R+F)$. When $R>0$ and $F=0$, JULIA reduces to the standard multi-linear model; when $R=0$ and $F>0$, JULIA reduces to the selected nonlinear tensor completion model.
		
		\item The selection of $R$ and $F$ plays a trade-off between multi-linear and nonlinear models, which can be used for JULIA to figure out the best combination of the numbers of multi-linear and nonlinear components, i.e., the ratio $R/F$.
	\end{itemize}
\end{remark}

\subsection{Alternating Optimization-based Initialization}
The optimization problem of JULIA is NP-hard \cite{hillar2013most}, so training such a mixture model is very challenging which requires very careful initialization. A na\"ive way to train JULIA is to employ sophisticated optimization methods such as Adam \cite{Kingma2015adam} or stochastic gradient descent (SGD) to learn all parameters at once. However, this is not the best way to train JULIA. We note that JULIA has two distinct components, and one can always fix the parameters in $g(\cdot)$ and optimize those in $f(\cdot)$, and vice versa. This naturally admits an alternating optimization (AO) design of the initial training process. Nevertheless, the na\"ive method ignores such a nice optimization structure. In the following, we present how to use AO to initialize our method.

For notation simplicity, in the following context, we define $\bTh_L=[\A_1,\cdots,\A_N]$ as the set that contains the multi-linear factor matrices and $\bTh_N=[\B_1,\cdots,\B_N,\bth]$ as the set that contains the nonlinear factor matrices and the associated parameter $\bth$ in $f(\cdot)$. With the above notations, we rewrite
\begin{align*}
	g\left(\{i_n,\A_n\}_{n=1}^N\right) &= g(\{i_n\}_{n=1}^N,\bTh_L) \\
	f\left(\{i_n,\B_n\}_{n=1}^N; \bth\right) &= f\left(\{i_n\}_{n=1}^N, \bTh_N\right).
\end{align*}
JULIA has the freedom of choosing the loss function for training, e.g., $t_1$-distance, Euclidean distance or Kullback–Leibler divergence, etc. As an example, let us consider the Euclidean distance as our loss function
\begin{align}\label{eq:lossfun}
	&L(\bTh_L, \bTh_N) = \sum_{i_1,\ldots,i_N\in\Omega}\notag\\ &\left( x_{i_1\ldots i_N} - g(\{i_n\}_{n=1}^N,\bTh_L) - f\left(\{i_n\}_{n=1}^N, \bTh_N\right) \right)^2
\end{align}
where $\Omega$ is the index set of the known tensor entries.

Then the optimization problem of JULIA becomes
\begin{align}\label{eq:optprob}
	\min_{\bTh_L,\bTh_N} L(\bTh_L, \bTh_N).
\end{align}
Here, it is obvious that $\bTh_L$ and $\bTh_N$ can be optimized alternately, i.e., we optimize $\bTh_N$ by fixing $\bTh_L$, and then we do the same for $\bTh_L$. 
More specifically, assuming that at the $t$-th iteration, there are some estimates of $\bTh_L$ and $\bTh_N$ available. Then at the $(t+1)$-th iteration, by given $\bTh_L^{(t)}$, the subproblem w.r.t. $\bTh_N$ is written as
\begin{align}
	\min_{\bTh_N} L(\bTh_L^{(t)}, \bTh_N)
\end{align}
This problem can be solved using first-order methods such as Adam \cite{Kingma2015adam}:
\begin{align}\label{eq:updatenonlinear}
	\bTh_N^{(t+1)} = \bTh_N^{(t)} - \mu_N\nabla_{\bTh_N} L(\bTh_L^{(t)}, \bTh_N^{(t)})
\end{align}
where $\mu_N$ is the learning rate of the nonlinear parameters and $\nabla_{\bTh_N} L(\bTh_L^{(t)}, \bTh_N)$ is the gradient of $L(\bTh_L^{(t)}, \bTh_N)$ w.r.t. $\bTh_N$ at the $(t+1)$-th iteration.

Similarly, by fixing $\bTh_N^{(t+1)}$, the update of $\bTh_L$ is given by
\begin{align}\label{eq:updatelienar}
	\bTh_L^{(t+1)} = \bTh_L^{(t)} - \mu_L\nabla_{\bTh_L} L(\bTh_L^{(t)}, \bTh_N^{(t+1)})
\end{align}
where $\mu_L$ is the learning rate of the multi-linear factor matrices.

\begin{remark}
	To start the JULIA algorithm, i.e., at $t=0$, we need an initial estimate of either $\bTh_L$ or $\bTh_N$. Since the estimation of $\bTh_L$ is a CP decomposition problem, it is relatively easier to be solved than the estimation of $\bTh_N$. Therefore, the initialization of JULIA is always recommended to start with $\bTh_L$. Given the number of multi-linear components, we first ignore the nonlinear parameters and initialize $\bTh_L$ by approximately solving the following problem with a fixed number of iterations:
	\begin{align*}
		\bTh_L = \arg\min_{\bTh_L} \sum_{i_1,\cdots,i_N\in\Omega}
		\left( x_{i_1\ldots i_N} - g\left(\{i_n,\A_n\}_{n=1}^N\right) \right)^2.
	\end{align*}
	Then we substitute $\bTh_L$ into \eqref{eq:lossfun} to estimate $\bTh_N$, and hence start the alternating optimization iterations. The detailed updating steps of AO initialization are summarized in Algorithm~\ref{alg:algorithm}.
\end{remark}

\begin{algorithm}[tb]
	\caption{AO Initialization}
	\label{alg:algorithm}
	\textbf{Input}: The observed tensor elements $\{x_{i_1\ldots l_N}\}$, $\forall i_1,\ldots,i_N\in\Omega$, the number of multi-linear components $R$ and the number of nonlinear components $F$\\
	\textbf{Parameter}: $\{\bTh_L,\bTh_N\}$\\
	\textbf{Output}: $\{x_{i_1\ldots l_N}\}$, $\forall i_1,\ldots,i_N\in\Omega^c$ where $\Omega^c$ denotes the complementary set of $\Omega$ that contains all indices of the missing values
	\begin{algorithmic}[1]
		\STATE Let $t=0$, and initialize the multi-linear factor matrices
		\WHILE{stopping criterion has not been reached}
		\STATE $t = t + 1$
		\STATE Update $\bTh_N$ via \eqref{eq:updatenonlinear}
		\STATE Update $\bTh_L$ via \eqref{eq:updatelienar}
		\ENDWHILE
	\end{algorithmic}
\end{algorithm}

\subsection{Training JULIA}
The training procedure of JULIA can be described in two steps:
\begin{enumerate}
	\item Employ Algorithm~\ref{alg:algorithm} to initialize the model parameters $\bTh_L$ and $\bTh_N$.
	\item Refine the model parameters by solving \eqref{eq:optprob} via Adam or SGD.
\end{enumerate}


\section{Experiments}

\subsection{Datasets}
We use six real-world tensors to examine the performance of the proposed method. The statistics of the six datasets are summarized in Table~\ref{tab:dataset}.
\begin{itemize}
	
	\item MovieLens (25M) \cite{harper2015movielens} is a dataset for movie recommendation which is structured as a 3-way tensor $\mathrm{user} \times \mathrm{movie} \times \mathrm{timestamp}$ in the following experiments. Each tensor element denotes a movie rate of movie $j$ given by user $i$ at timestamp $k$. Note that we only select users with at least 20 ratings in the dataset. For users with more than 3000 ratings, we only keep the first 3000 ratings according to their respective timestamps in ascending order. 
	
	\item Facebook Wall Posts \cite{rossi2015network} is a dataset that collects the number of wall posts from one Facebook user to another over a period of 1506 days. 
	We transform the timestamp to date and create a $\mathrm{user post}\times\mathrm{user wall}\times\mathrm{date}$ tensor via grouping by $\mathrm{user post}$, $\mathrm{user wall}$ and $\mathrm{date}$ and calculate the daily number of posts from user-$i$ to user-$j$'s wall.
	
	\item Healthcare is a spatio-temporal dataset that records the daily number of medical claims of various diseases at a county level in the United States. There are 2976 counties, 282 diseases, 202 medical procedures, and 728 days from 2018-12-29 to 2020-12-25.
	
	\item Enron-Emails \cite{shetty2004enron} was released during an investigation by the Federal Energy Regulatory Commission. The modes represent $\mathrm{sender}\times\mathrm{receiver}\times\mathrm{word}\times\mathrm{date}$, and the values are counts of words. 
	
	\item NeurIPS publication \cite{chechik2007eec} collects the papers published in NIPS from 1987 to 2003. The modes represent $\mathrm{paper}\times\mathrm{author}\times\mathrm{word}\times\mathrm{year}$, and the values are counts of words.
	
	\item Uber-Pickups consists of six months of Uber pickup data in New York City during April 2014 to August 2014, provided by fivethirtyeight\footnote{\url{https://www.kaggle.com/fivethirtyeight/uber-pickups-in-new-york-city}} after a Freedom of Information request. The modes represent $\mathrm{dates}\times\mathrm{hours}\times\mathrm{latitudes}\times\mathrm{longitudes}$, and the values are number of pickups.
\end{itemize}

\begin{table}[t]
	\centering
	\resizebox{\columnwidth}{!}{
		\begin{tabular}{lll}
			\toprule
			Dataset & Shape & \# known \\
			\midrule
			MovieLens & (139357, 57675, 2981) & 20260421 \\
			Facebook & (42390, 39986, 1506) & 738078 \\
			Healthcare & (2814, 266, 189, 728) & 9002336 \\
			Enron-Emails & (6066, 5699, 244268, 1176) & 54202099 \\
			NeurIPS & (2482, 2862, 14036, 17) & 3101609 \\
			Uber-Pickups & (183, 24, 1140, 1717) & 3309490 \\
			\bottomrule
		\end{tabular}
	}
	\caption{Statistics of real-world tensors.}
	\label{tab:dataset}
\end{table}

\subsection{Baseline methods}

We compare the proposed method to the state-of-the-art tensor completion algorithms, including both multi-linear and nonlinear models. The baseline methods are summarized as follows:
\begin{itemize}
	\item COSTCO \cite{liu2019costco} is based on the convolutional neural network that models nonlinear interactions between tensor elements while preserving the low-rank structure.
	
	\item AVOCADO \cite{schreiber2020avocado} is a deep neural network model that concatenates all embedding vectors into a wide vector which is then passed through an MLP.
	
	\item NeurTN \cite{chen2020learning} is a neural powered Tucker network model.
	
	\item Tucker completion is the standard Tucker tensor completion algorithm solved using stochastic gradient descent.
	
	\item CP completion (CPC) is the standard multi-linear low-rank tensor completion method solved using stochastic gradient descent.
\end{itemize}

Note that we test JULIA with different combinations of $R$ and $F$. We name JULIA as ``JULIA ($R/F$)'', e.g., ``JULIA ($3/7$)'' means $R=3$ multi-linear components and $F=7$ nonlinear components where the tensor rank is $R+F=10$. Throughout of the examples, the learning rate is $l_r = 0.005$. For the baseline methods, we employ Adam \cite{Kingma2015adam} for training. Unless otherwise emphasized, the total number of latent components in all methods except NeurTN is 20. The NeurTN method does not work well with rank 20 in the testing datasets. Instead of 20, we choose rank 10 for this method which works much better.
We employ a 3-layer MLP in JULIA, AVOCADO, NeurTN, and AVOCADO, where the layer dimension for JULIA is $\{NR\times F^2, F^2\times F, F\times 1\}$ and that for the remaining methods is $\{N(R+F)\times N^2(R+F), N^2(R+F)\times (R+F), (R+F)\times 1\}$. All experiments were run through a platform with 4 CPUs, 128 GB RAM, and one Nvidia V100 GPU.

\subsection{Metrics}
The metrics for performance comparison includes the root mean-squared-error (RMSE), mean-absolute-error (MAE) and relative fitting error (RFE), where the RE is computed as
$$\mathrm{RFE} = \left\|\mathcal{P}_{\Omega^c} (\hat{\tX} - \tX)\right\|_F / \left\|\mathcal{P}_{\Omega^c} (\tX)\right\|_F
$$
where $\mathcal{P}_{\Omega^c}$ represents the mask of the missing entries.

In the following experiments, for each dataset, 80\% of the data points are used for training while the remaining 20\% for testing. We further randomly select 10\% of the training data as a validation set for the early stopping purpose, where the stopping criterion is the relative validation error between two adjacent epochs below a threshold $10^{-4}$. We observe that COSTCO and NeurTN can fail to converge when the dataset is large but has a limited number of data points, resulting in poor testing performance. To overcome this issue, whenever we observe a failure, i.e., $\mathrm{RFE}\geq 1$, we restart them with a new random initialization until the total number of restarts reaches 10. 

\begin{figure}[t]
	\centering
	\includegraphics[width=\columnwidth]{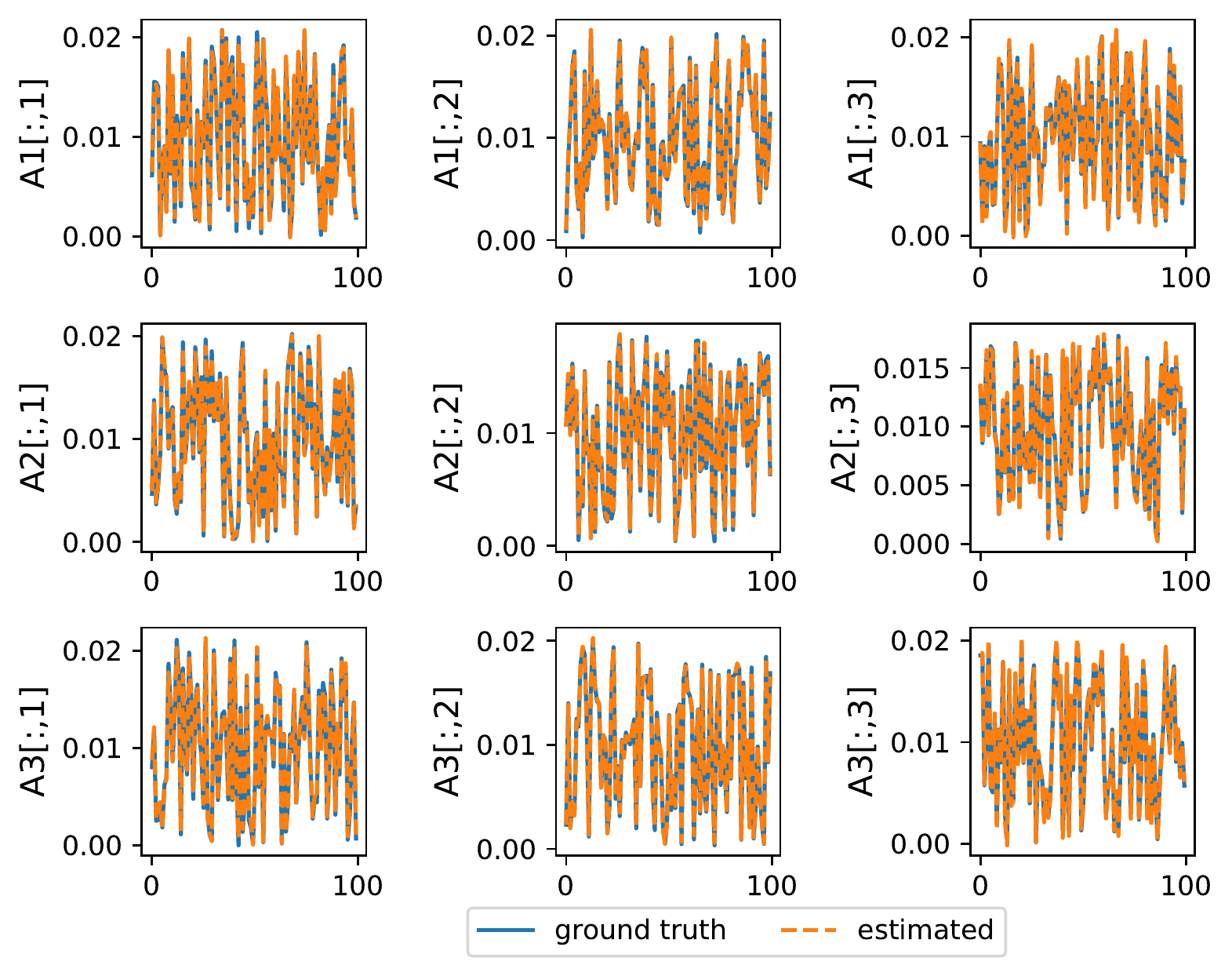}
	\caption{Example of JULIA identifying multi-linear components from a mixture of multi-linear and nonlinear components. Its RMSE on the testing set reaches $0.0012$ after 11 iterations. Components are normalized and aligned using the Hungarian algorithm \cite{kuhn1955hungarian}.}
	\label{fig:identifiability}
\end{figure}

\subsection{Results}

\subsubsection{Identifiability of JULIA}

We first study JULIA's capability of identifying the multi-linear components in tensors. To examine the identifiability, we generate a $100\times100\times100$ synthetic tensor in which there are $R=3$ multi-linear components, $F=10$ nonlinear components, and 80\% of the tensor elements are missing. We split the observed data into 80\% for training and 20\% for testing. We employ JULIA with $R=3$ and $F=10$ to estimate the multi-linear components. Figure~\ref{fig:identifiability} shows the results, where we see that all the estimated latent components are identical to the ground truth one, which verifies its capability of uniquely recovering the multi-linear components.



\begin{figure}[t]
	\centering
	\subfigure[Varying rank with fixed $R/F$ ratio]{\includegraphics[width=1\columnwidth]{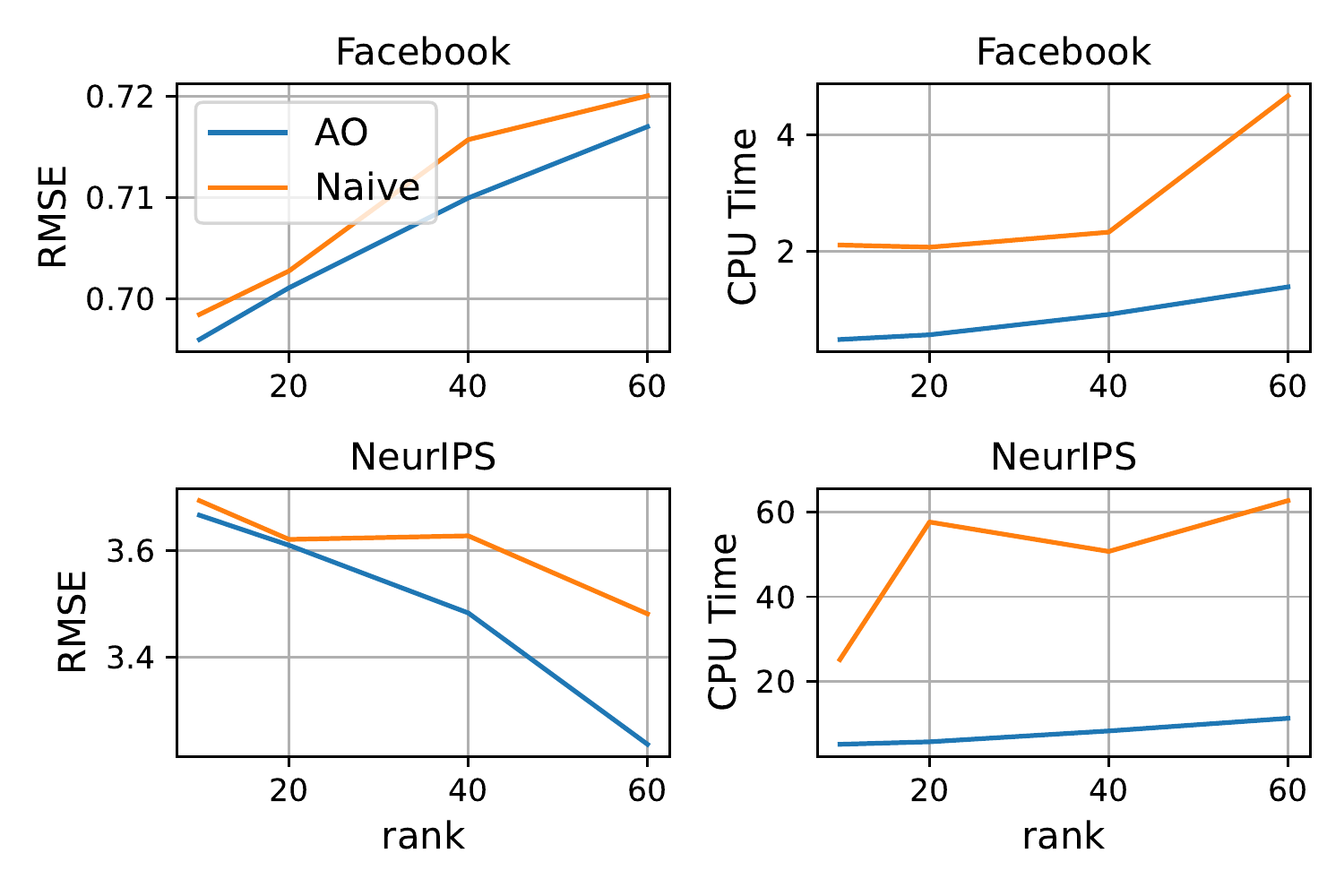}\label{fig:juliarank}}
	\subfigure[Varying $R/F$ ratio with fixed rank]{\includegraphics[width=1\columnwidth]{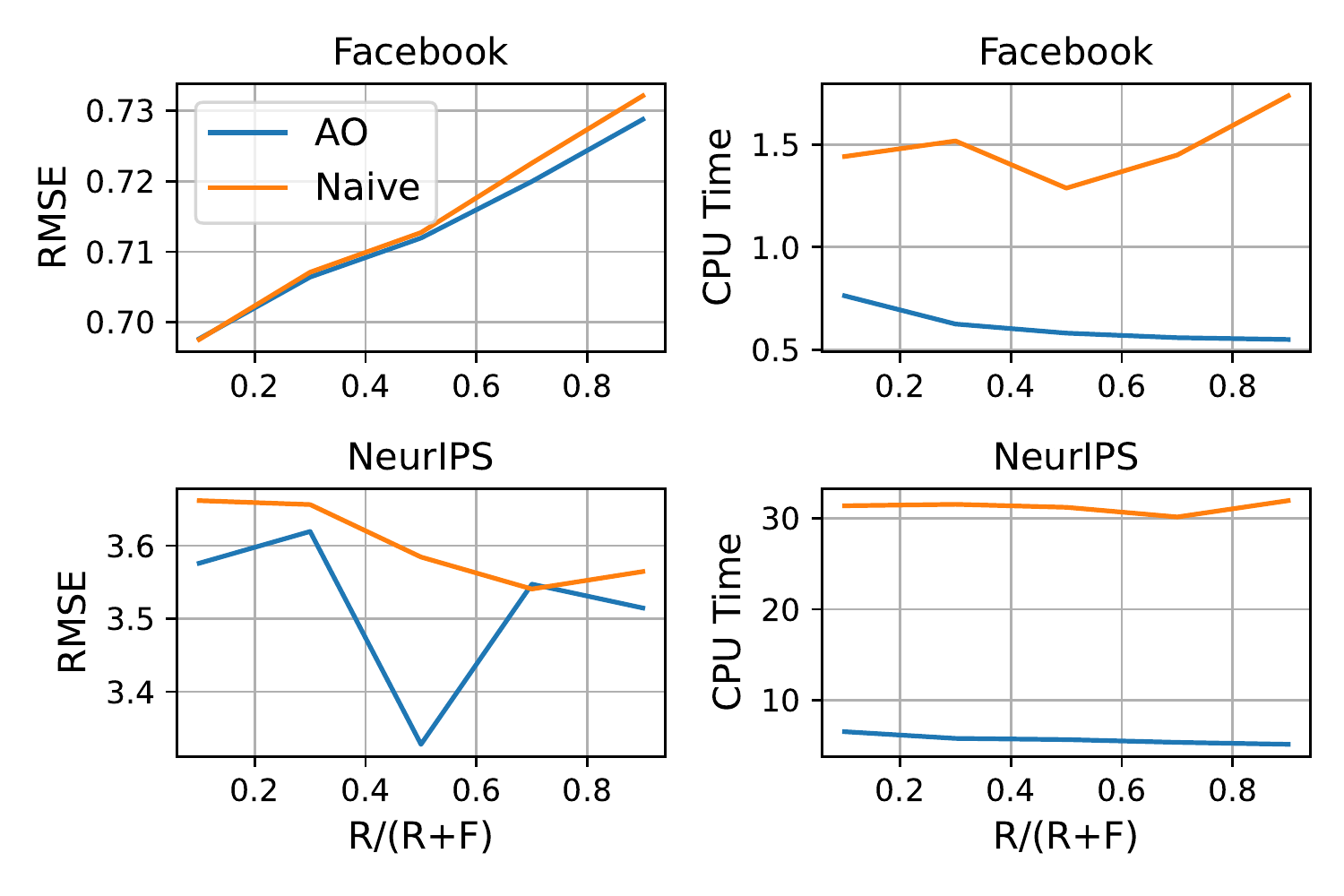}\label{fig:juliaratio}}
	\caption{AO vs Na\"ive initialization}
	\label{fig:juliahyper}
\end{figure}

\subsubsection{AO vs. na\"ive training for JULIA} 
This example shows that  JULIA with the AO initialization works better than the na\"ive training method. We increase the number of components (i.e., rank) from 10 to 60 and for each rank and set 20\% as multi-linear components and the remaining 80\% as nonlinear components. We compare the performance of JULIA initialized with AO and na\"ive random initializations, respectively. The maximum number of AO iteration is 20. Figure~\ref{fig:juliahyper} shows that JULIA with AO outperforms the na\"ive random initialization consistently in the Facebook and neurips datasets with higher accuracy and less time consumption. The CPU time improvement implies that AO can provide better initialization to accelerate the subsequent optimization. 

\begin{table*}[htbp]
	\centering
	\begin{tabular}{lllllllllllll}
		\toprule
		& \multicolumn{2}{c}{MovieLens} & \multicolumn{2}{c}{Facebook} & \multicolumn{2}{c}{Healthcare} & \multicolumn{2}{c}{Enron-Emails} & \multicolumn{2}{c}{NeurIPS} & \multicolumn{2}{c}{Uber-Pickups} \\
		\cmidrule{2-13}    Method & RSE   & RMSE  & RSE   & RMSE  & RSE   & RMSE  & RSE   & RMSE  & RSE   & RMSE  & RSE   & RMSE \\
		\midrule
		JULIA & 0.225 & 0.825 & \textbf{0.506} & \textbf{0.701} & \textbf{0.330} & \textbf{35.746} & \textbf{0.363} & \textbf{4.839} & \textbf{0.736} & \textbf{3.509} & \textbf{0.466} & \textbf{0.795} \\
		COSTCO & 0.232 & 0.850 & 0.515 & 0.714 & 0.499 & 54.034 & 0.364 & 4.856 & 0.790 & 3.770 & 0.486 & 0.830 \\
		AVOCADO & \textbf{0.223} & \textbf{0.817} & 0.530 & 0.739 & \textbf{0.330} & 35.757 & 0.455 & 5.974 & 0.787 & 3.733 & 0.513 & 0.875 \\
		NeurTN & 0.224 & 0.820 & 0.508 & 0.704 & 0.340 & 36.799 & 1.000 & 13.337 & 0.773 & 3.689 & 1.000 & 1.707 \\
		CPC   & 0.237 & 0.870 & 0.552 & 0.766 & 0.447 & 48.416 & 0.438 & 5.837 & 0.781 & 3.724 & 0.491 & 0.838 \\
		Tucker & 0.240 & 0.880 & 0.973 & 1.357 & 1.000 & 108.306 & 0.618 & 8.248 & 0.811 & 3.870 & 1.015 & 1.732 \\
		\bottomrule
	\end{tabular}
	\caption{Performance comparison of JULIA with baseline methods.}
	\label{tab:juliacomparebaseline}%
\end{table*}%

\begin{table*}[t]
	\centering
	\resizebox{\linewidth}{!}{
		\begin{tabular}{lllllllllllll}
			\toprule
			& \multicolumn{2}{c}{MovieLens} & \multicolumn{2}{c}{Facebook} & \multicolumn{2}{c}{Healthcare} & \multicolumn{2}{c}{Enron-Emails} & \multicolumn{2}{c}{NeurIPS} & \multicolumn{2}{c}{Uber-Pickups} \\
			\cmidrule{2-13}    Method & RSE   & RMSE  & RSE   & RMSE  & RSE   & RMSE  & RSE   & RMSE  & RSE   & RMSE  & RSE   & RMSE \\
			\midrule
			JULIA-COSTCO (4/16) & 0.226 & 0.829 & \textbf{0.508} & \textbf{0.704} & \textbf{0.361} & \textbf{39.080} & 0.483 & 6.437 & \textbf{0.636} & \textbf{3.036} & \textbf{0.468} & \textbf{0.799} \\
			JULIA-COSTCO (10/10) & \textbf{0.216} & \textbf{0.792} & 0.515 & 0.714 & 0.468 & 50.643 & 0.397 & 5.300 & 0.671 & 3.202 & 0.477 & 0.814 \\
			JULIA-COSTCO (16/4) & 0.218 & 0.799 & 0.523 & 0.726 & 0.431 & 46.615 & 0.412 & 5.501 & 0.701 & 3.346 & 0.477 & 0.814 \\
			COSTCO & 0.232 & 0.850 & 0.515 & 0.714 & 0.499 & 54.034 & \textbf{0.364} & \textbf{4.856} & 0.790 & 3.770 & 0.486 & 0.830 \\
			\midrule
			\midrule
			JULIA-AVOCADO (4/16) & 0.225 & 0.823 & \textbf{0.508} & \textbf{0.704} & \textbf{0.308} & \textbf{33.302} & 0.550 & 7.343 & 0.768 & 3.666 & \textbf{0.466} & \textbf{0.795} \\
			JULIA-AVOCADO (10/10) & \textbf{0.218} & \textbf{0.799} & 0.514 & 0.713 & 0.334 & 36.122 & 0.470 & 6.269 & \textbf{0.710} & \textbf{3.388} & 0.489 & 0.834 \\
			JULIA-AVOCADO (16/4) & 0.221 & 0.809 & 0.526 & 0.729 & 0.432 & 46.744 & \textbf{0.426} & \textbf{5.680} & 0.756 & 3.609 & 0.479 & 0.818 \\
			AVOCADO & 0.223 & 0.817 & 0.530 & 0.739 & 0.330 & 35.757 & 0.455 & 5.974 & 0.787 & 3.733 & 0.513 & 0.875 \\
			\midrule
			\midrule
			JULIA-NeurTN (2/8) & 0.224 & 0.819 & \textbf{0.507} & \textbf{0.703} & \textbf{0.316} & \textbf{34.249} & 0.643 & 8.576 & \textbf{0.770} & \textbf{3.675} & 0.519 & 0.885 \\
			JULIA-NeurTN (5/5) & 0.218 & 0.799 & 0.510 & 0.707 & 0.472 & 51.087 & 0.447 & 5.961 & 0.772 & 3.681 & 0.499 & 0.852 \\
			JULIA-NeurTN (8/2) & \textbf{0.217} & \textbf{0.794} & 0.511 & 0.709 & 0.438 & 47.441 & \textbf{0.445} & \textbf{5.941} & 0.754 & 3.595 & \textbf{0.491} & \textbf{0.838} \\
			NeurTN & 0.224 & 0.820 & 0.508 & 0.704 & 0.340 & 36.799 & 1.000 & 13.337 & 0.773 & 3.689 & 1.000 & 1.707 \\
			\bottomrule
		\end{tabular}
	}
	\caption{Performance comparison of baseline methods with or without JULIA.}
	\label{tab:baselinejulia}%
\end{table*}

\subsubsection{Performance comparison with baseline methods}

\begin{figure}
	\centering
	\includegraphics[width=1\columnwidth, trim={0, 0.5cm, 0, 0}, clip]{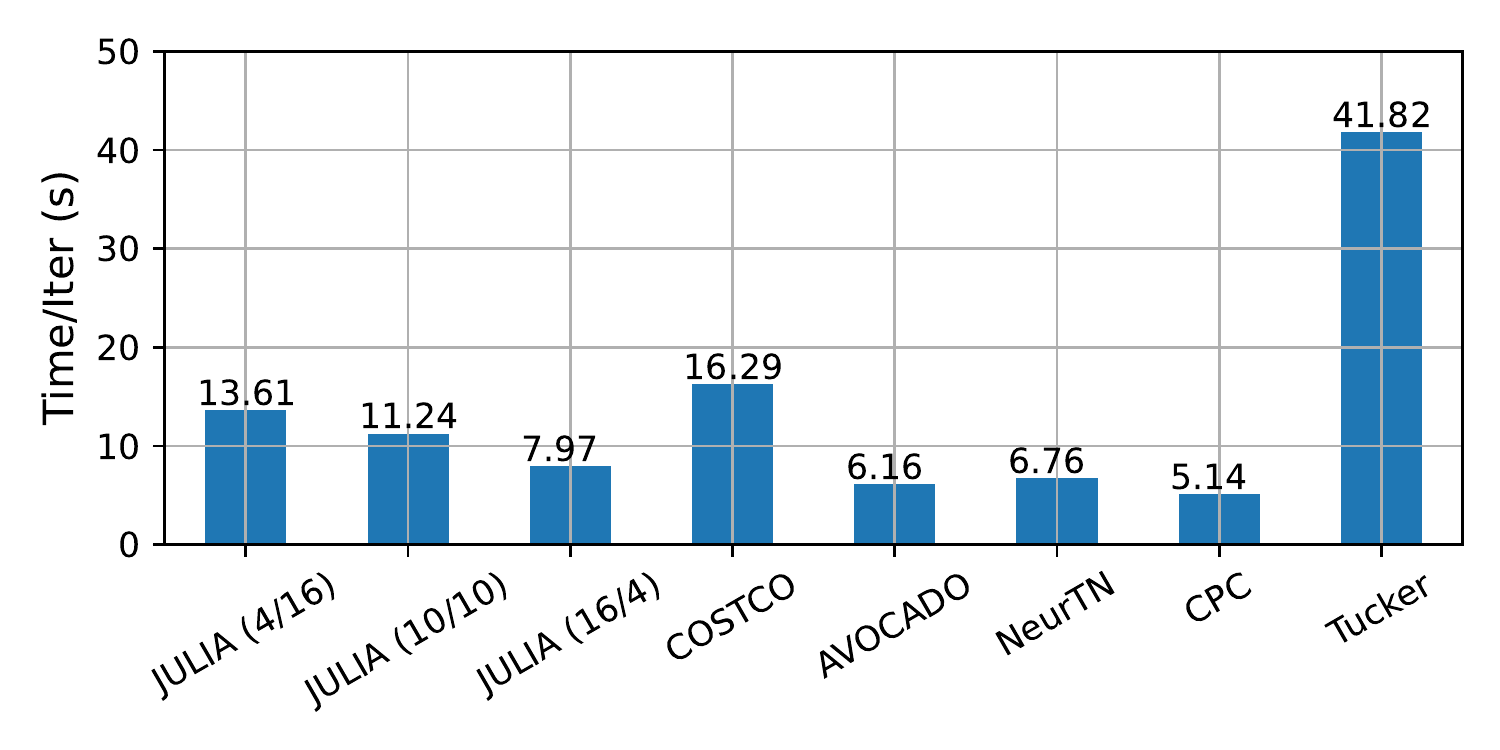}
	\caption{Time complexity comparison on the Enron-Emails dataset.}
	\label{fig:time}
\end{figure}

\begin{figure}[t]
	\centering
	\includegraphics[width=1\linewidth, trim={0.2cm 0.5cm 0cm 0.3cm},clip]{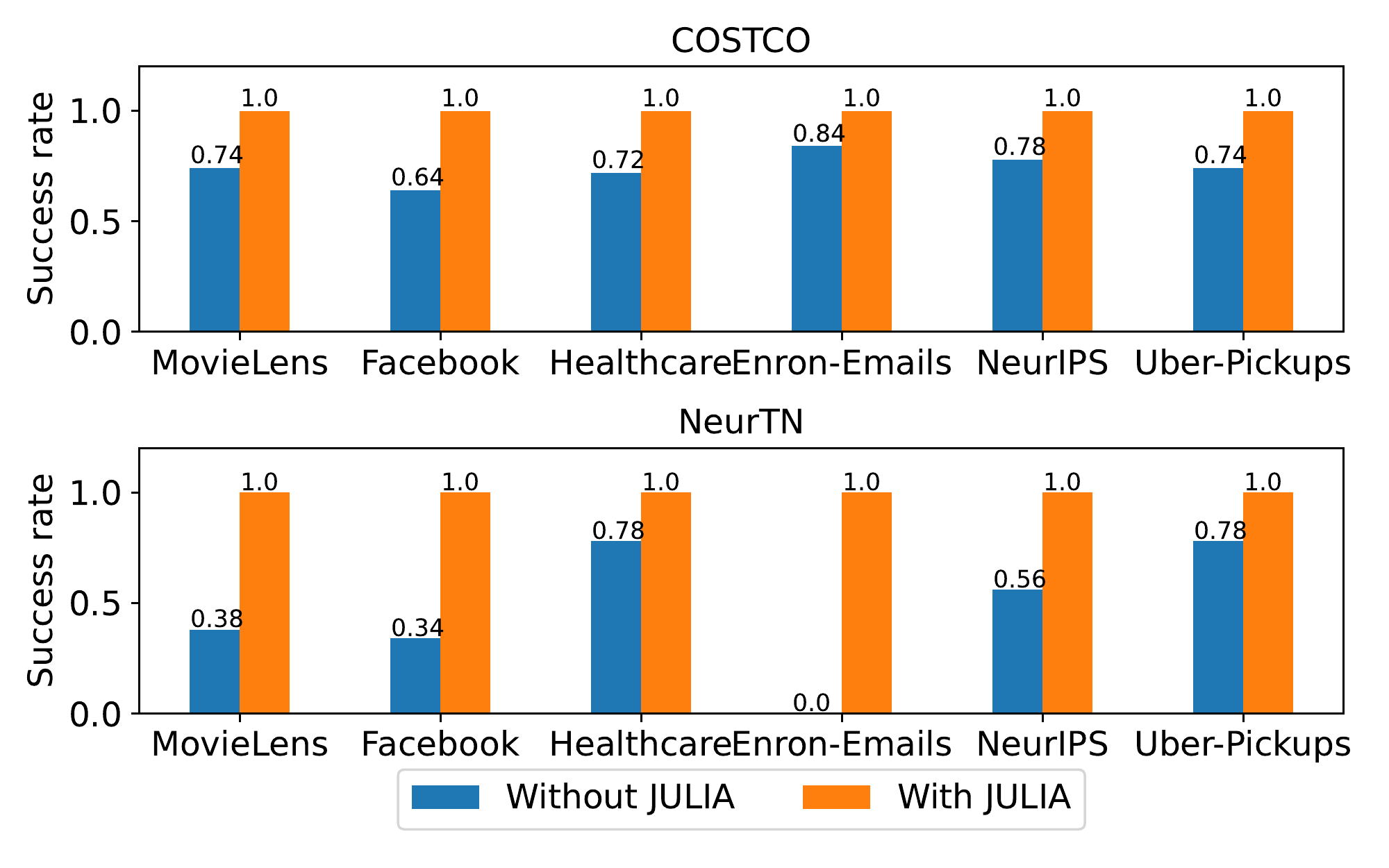}
	\caption{Success rate comparison of methods with or without the JULIA framework, where $R=4$ and $F=16$.}
	\label{fig:succrate}
\end{figure}

We now compare JULIA with five baseline methods. Table~\ref{tab:juliacomparebaseline} shows the results. In the Facebook, Healthcare, Enron-Emails, NeurIPS, and Uber-Pickups datasets, JULIA has the smallest RFE and RMSE. In the MovieLens dataset, AVOCADO performs the best, but its accuracy is very close to that of the JULIA and NeurTN methods. The NeurTN method fails to work in the Enron-Emails and Uber datasets. One reason could be that NeurTN is over parameterized than the other neural network methods.

Figure~\ref{fig:time} shows the time complexity comparison with the largest dataset Enron-Emails, where the CPU time per iteration (epoch) has been plotted. The time complexity of JULIA decreases as $R/F$ increases. This is because given a tensor rank, the total number of components is deterministic, and a smaller $F$ means fewer parameters in $\bTh_N$. The time complexity of JULIA (16/4) is comparable to AVOCADO and NeurTN, but is only half of the complexity of COSTCO. Tucker takes the longest time is due to its efficiency in estimating the core-tensor in a large-scale 4-way tensor.

\subsubsection{Incorporating JULIA into existing methods}
We incorporate JULIA into the existing tensor completion methods to examine if JULIA can improve its performance. We consider COSTCO, AVOCADO, and NeurTN as examples and name them JULIA-COSTCO, JULIA-AVOCADO, and JULIA-NeurTN, respectively. Specifically, we replace the nonlinear function $f(\cdot)$ in JULIA with each of the three algorithms and use the developed alternating optimization method to train the method. Since JULIA splits the tensor rank into linear and nonlinear parts, we consider three cases to evaluate the performance, i.e., $(R=4,F=16)$, $(R=10,F=10)$ and $(R=16,F=4)$, where the tensor rank is 20. The results are shown in Table~\ref{tab:baselinejulia}. Except for the Enron-Emails dataset, all methods with JULIA achieve significant performance improvement and outperform their respective `na\"ive' versions. COSTCO with JULIA obtains 27.7\% and 19.5\% lower RMSE compared to its na\"ive implementation in the Healthcare and NeurIPS datasets, respectively. Note that the NeurTN method after using JULIA achieves the biggest improvement. It outperforms the na\"ive NeurTN method significantly and achieves more than 50\% RMSE improvement in the Enron-Emails and Uber-Pickups datasets.

\begin{table}[htbp]
	\centering
	\resizebox{\linewidth}{!}{
		\begin{tabular}{c|lccc}
			\toprule
			& Method & AVOCADO & COSTCO & NeurTN \\
			\midrule
			\multirow{4}{*}{\shortstack{Total \\ Time (s)}} & Naive & 486.2 & 4793.9 & 583.0 \\
			& JULIA (4/16) & 762.4 & 1545.2 & 891.2 \\
			& JULIA (10/10) & 359.1 & \textbf{569.7} & 210.5 \\
			& JULIA (16/4) & \textbf{97.7} & 784.5 & \textbf{192.7} \\
			\midrule
			\multirow{4}{*}{\shortstack{Time per \\ Iter. (s)}} & Naive & 8.4   & 33.8  & 6.6 \\
			& JULIA (4/16) & 8.3   & 24.9  & 7.7 \\
			& JULIA (10/10) & 6.2   & 16.3  & 5.0 \\
			& JULIA (16/4) & \textbf{4.1} & \textbf{10.5} & \textbf{3.7} \\
			\bottomrule
		\end{tabular}%
	}
	\caption{Time complexity comparison between existing methods and their JULIA versions.}
	\label{tab:timebaseline}%
\end{table}%

Although JULIA does not improve the accuracy of COSTCO in the Enron-Emails dataset, it is still worth mentioning that JULIA can improve the convergence of COSTCO, hence reducing its time complexity. Table~\ref{tab:timebaseline} verifies such an observation, where after applying JULIA, the total running time of AVOCADO, COSTCO, and NeurTN have been reduced by 77.9\%, 84\%, and 66.9\%, respectively, while their time per iteration has been reduced by 51.2\%, 68.9\%, and 43.9\%, respectively.
Furthermore, Figure~\ref{fig:succrate} shows the success rates of JULIA, NeurTN, and their JULIA variants, where The $\mathrm{success~rate}$ is defined as
$$
\mathrm{success~rate} = \frac{\#~of~\mathrm{successful~tests}}{\#\mathrm{~of~tests~in~total}}.
$$

In this example, we test each method using 50 independent tests and record their RFE values. We define a method that is successful if and only if its RFE is smaller than one, otherwise, unsuccessful. 
It is seen in Fig.~\ref{fig:succrate} that the $\mathrm{success~rate}$ of COSTCO and NeurTN are lower than 1, meaning that they cannot guarantee to solve the tensor completion accurately. For COSTCO, its $\mathrm{success~rate}$ is around 0.74 in most datasets. NeurTN has much worse $\mathrm{success~rate}$, and for the Enron-Emails dataset, it never achieves a satisfactory result and which results in large RFE and RMSE in Table~\ref{tab:juliacomparebaseline}. Compared to the `vanilla' COSTCO and NeurTN, after implementing JULIA, their $\mathrm{success~rate}$s increase to 100\%, and their RFE and RMSE are improved significantly as well, see Table~\ref{tab:baselinejulia}.

\section{Conclusion}
In this paper, we propose JULIA as a general framework for large-scale tensor completion. JULIA models tensors using two networks: a multi-linear network learning multi-linear latent components and a nonlinear DNN learning nonlinear latent components from incomplete tensors, such that every tensor element is represented by a sum of multi-linear and nonlinear estimates. Experimental results have shown that JULIA is more efficient and accurate in solving the tensor completion problem than many recently developed SOTA methods. Furthermore, we show that JULIA can be implemented to improve the performance of existing tensor completion methods and increase their success rate.

\bibliographystyle{ieeetran}
\bibliography{IEEEabrv, my}

\end{document}

%% file: def.tex
\def\bth{\boldsymbol{\theta}}

\def\bTh{\mathbf{\Theta}}

\def\a{\mathbf{a}}
\def\A{\mathbf{A}}

\def\b{\mathbf{b}}
\def\B{\mathbf{B}}

\def\bR{\mathbb{R}}

\def\w{\mathbf{w}}

\def\X{\mathbf{X}}

\def\z{\mathbf{z}}

\def\tX{\underline{\X}} 

\newtheorem{remark}{Remark}